# Haptic-Enhanced Virtual Reality Simulator for Robot-Assisted Femur Fracture Surgery*

Fayez H. Alruwaili, David W. Halim-Banoub, Jessica Rodgers, Adam Dalkilic, Christopher Haydel, Javad Parvizi, Iulian I. Iordachita, Mohammad H. Abedin-Nasab

*Abstract*— In this paper, we develop a virtual reality (VR) simulator for the Robossis robot-assisted femur fracture surgery. Due to the steep learning curve for such procedures, a VR simulator is essential for training surgeon(s) and staff. The Robossis Surgical Simulator (RSS) is designed to immerse user(s) in a realistic surgery setting using the Robossis system as completed in a previous real-world cadaveric procedure. The RSS is designed to interface the Sigma-7 Haptic Controller with the Robossis Surgical Robot (RSR) and the Meta Quest VR headset. Results show that the RSR follows user commands in 6 DOF and prevents the overlapping of bone segments. This development demonstrates promising avenue for future implementation of the Robossis system.

## I. INTRODUCTION

Surgical competence is achieved through hours of practice and failure using models under the supervision of a limited number of highly specialized surgeons [1]. This can be very time-consuming and costly for future specialized surgeons to attain the experience needed for operational proficiency. Virtual reality (VR) simulators allow residents and skilled surgeons to learn new complex surgical procedures through failure with low risk [2]. Through VR training, surgeons have demonstrated shortened surgical times, greater tool dexterity, and greater accuracy in the operating room [3]. When used for training, VR simulators equipped with haptic feedback have demonstrated improved skill acquisition in complex surgeries, including but not limited to laparoscopy, endoscopy, cataract surgery, and neurosurgery [3], [4].

Currently, long bone femur fracture surgery has a high risk of surgical complications, including high malalignment rates and high fracture reduction forces. In addition, the exposure to X-ray radiation and extended amounts of time in operation can harm patients of this surgery [5]–[8]. To reduce complication risk, our group presented a surgical system called Robossis that aids in eliminating avoidable complications [5]–[8]. Robossis has shown the potential to eliminate these complications through cadaveric and benchtop studies, but user training is required to maximize fluidity and success rate [5]–[8].

## II. RELATED WORK AND CONTRIBUTION

The number of robot-assisted surgeries continues to grow annually, and the training of surgeon(s) and operating staff to utilize these devices effectively has been studied. In the past, a variety of surgical simulators, including RoSS, dV-Trainer, dVSS, and SEP, were developed to provide surgeon(s) and operating staff with the required skills for the varying surgical robotic systems [9]–[12]. Further, multiple studies validated the effectiveness of the training regime using the surgical simulators, which shows a significant improvement in surgical proficiency translated to the operating room [9]–[12].

As such, we aim to create the 1$^{st}$ Robossis Surgical Simulator (RSS) that is designed specifically for femur fracture surgeries. We aim to provide the surgeon(s) and operating staff with the required, accurate training resources for the Robossis system. The RSS is developed to immerse the users in a 3D environment utilizing the Meta Quest VR headset (Meta – United States) and haptic feedback via the Sigma.7 haptic controller (Force Dimension – Switzerland). Furthermore, we leverage Unreal Engine with high-end graphics and advanced rendering capabilities for creating a high-quality VR environment. The key aspects of our design and development in this paper is the following:

1. We design and develop the RSS that inherits a surgical environment as previously completed in a cadaver experiment. Also, we design a control architecture that integrates the user to the VR environment using the HC and Meta Quest VR headset. Further, we leverage the tools of unreal engines to provide 2D fluoroscopic imaging within the VR environment.
2. We develop the kinematic representation of the Robossis Surgical Robot (RSR) and Sigma.7 Haptic Controller (HC) within the surgical simulator. We implement a motion controller to drive the joints of the RSR and HC as resembled in the real world. We validate the kinematic representation by performing simulation error analysis.
3. We develop a force feedback collision algorithm that projects forces into the user's hand via the HC to prevent

*This work is funded by the National Science Foundation (NSF) under grants 2141099, and 2226489, and by the New Jersey Health Foundation (NJHF) under grant PC 62-21.

F. A, A. D, and M. A are with the Biomedical Engineering Department, Rowan University, Glassboro, NJ 08028, USA (e-mail: alruwa16@rowan.edu, dalkil38@students.rowan.edu, corresponding author's email: abedin@rowan.edu).

D. H is with Rowan-Virtua School of Osteopathic Medicine, Stratford, NJ 08084, USA(e-mail: halimb73@rowan.edu)

J. R is with the Mechanical Engineering, Rowan University, Glassboro, NJ 08028, USA (e-mail: rodger58@students.rowan.edu ).

C. Haydel is an Orthopedic Trauma Surgery with Virtua Health, Moorestown, NJ 08057 (e-mail: chaydel@virtua.org)

J. Parvizi is with Rothman Orthopedic Institute, Thomas Jefferson University Hospital, Philadelphia, Pennsylvania (e-mail: javadparvizi@gmail.com)

I. I. is with the Laboratory for Computational Sensing and Robotics, Johns Hopkins University, Baltimore, MD 21218 USA (e-mail: iordachita@jhu.edu).

an overlap between the proximal and distal bone. We model the proximal and distal bones as an oriented bounding box (OBB) and retrieve the collision utilizing the separating axis theorem. Thus, 4 OBBs are designed to cover the shaft, distal, and proximal segments of the femur bone for realistic real-life modeling.

## III. ARCHITECTURE OF THE ROBOSSIS SURGICAL SIMULATOR

The architecture of the RSS is illustrated in Fig. 1. The VR environment was designed using Unreal Engine 5.2.1 and Blender 3.6. We use the HC as the interface between the user input trajectories ($X_{HC}$) and speed ($\dot{X}_{HC}$) into the VR environment to the manipulation the RSR. We implement a motion control algorithm that scales user input trajectories to a maximum linear and angular velocity to represent similar conditions to the real world. Also, we develop the kinematic representation of the RSR and HC to resemble real-world physical systems. We determine the inverse kinematics of the RSR and HC to drive the joints of the RSR ($d_{RSR,i}, \theta_{RSR,i}$) and HC ($\theta_{HC,i}$) and manipulate the end-effector of each robot to the desired location and orientation within the RSS. Also, we implement a haptic feedback algorithm ($F_{col}$) that restricts the overlap between the distal (D) and proximal (P) bone segment ($D \cap P$). Additionally, we incorporate the Meta Quest VR headset to immerse the user into a 3D virtual environment by using the Oculus VR plugin impeded within the unreal engine. Also, we utilize the input from the Oculus headset to provide the user with additional control over the orientation of the c-arm to capture the 2D fluoroscopic imaging.

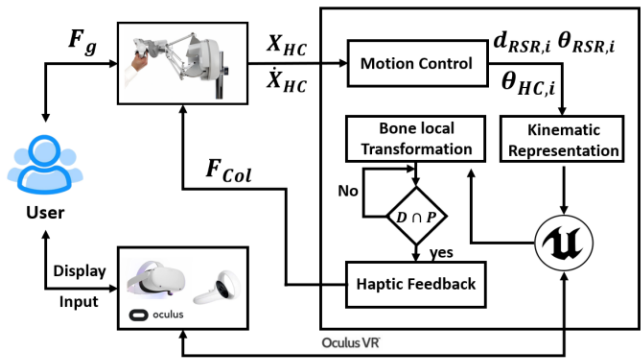

Figure 1. The architecture of the designed RSS is illustrated, where the entire architecture includes the haptic Sigma-7 controller, the Metal Quest VR headset, control algorithms, kinematic representation, haptic feedback, and the Unreal Engine simulator to house the VR environment.

## IV. SIMULATOR DESIGN AND MODELING

The RSS was modeled to resemble an actual operating room for the Robossis system surgical setting for femur fracture surgery, as previously completed in a cadaver experiment (Fig. 2) [8].

### A. Surgical Environment Design

The RSS was designed using Unreal Engine 5.2.1 and Blender 3.6. The RSS includes the HC, a surgeon workstation, RSR, the patient placed in the supine position, and the C-arm X-ray machine. Also, the RSR is attached to the patient's distal femur using surgical rods. Blender software was used to provide the required enhancement for the visual rendering of the meshes. For example, Blender was used to provide draping of the patient, coloring of each meshes, and establishing the reference frame for the translation and orientation of the meshes. Furthermore, the RSS, inherited from a VR template, was designed to interface the environment with external hardware control algorithms and house the surgical simulator. Additionally, the simulation was designed for integration with the Meta Quest headset to establish an immersive and in-depth VR environment. The Meta Quest controller was integrated to facilitate secondary simulation controls. This controller takes user input to direct the C-Arm X-ray position and rotation, enabling various anatomical planar views of the surgical field, which are consistently updated, producing a real-time display X-ray imaging monitor.

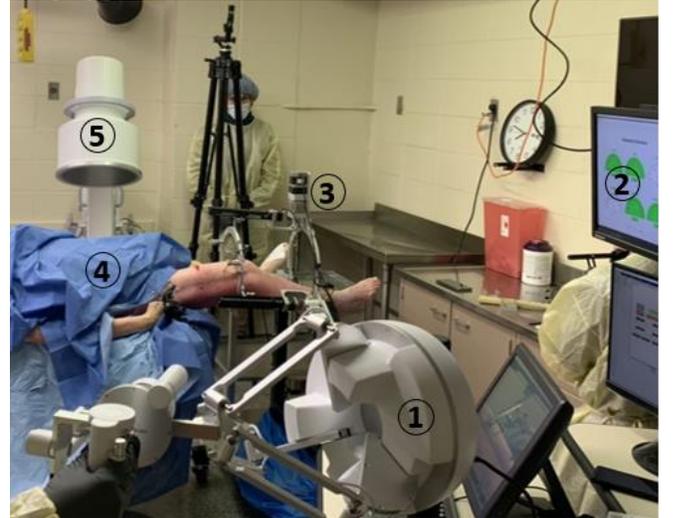

Figure 2. A previous cadaver experiment using the Robossis System. The surgical setting includes (1) a haptic controller, (2) a surgeon workstation, (3) the Robossis Surgical robot, (4) a cadaver patient, and (5) a C-arm X-ray machine.

### B. Robot Kinematic Representation

Robossis system consists of a leader HC Sigma.7 and a follower RSR. The sigma.7 HC is a hybrid robot structure based on a delta mechanism providing 3-DOF translational manipulation, a wrist serial mechanism providing 3-DOF rotational manipulation, and a grasping unit for 1-DOF (Fig. 3A). To define the kinematic representation of the HC in the unreal engine, the HC components were divided into varying links ($L_{HC}$1-9) and connected via joints ($J_{HC}.A_i$-$H_i$) using the parent-child convention to define the relationship between the links (Fig. 3A). For each arm of the delta mechanism, it consists of one active joint ($J_{HC}.A_i$) and six passive joints ($J_{HC}.C_i$-$E_i$). Further, each arm is connected to a fixed base ($L_{HC}$6) connected to the serial wrist mechanism. The wrist serial mechanism consists of three active joints ($J_{HC}.F_i$-$H_i$) responsible for the three independent axes of rotation ($\alpha$ $\beta$ $\gamma$). As such, a closed-loop link-joint relationship is defined to resemble the actual real-world equivalence of the Sigma.7 HC.

Additionally, the follower RSR is a 3-armed parallel mechanism where each arm is placed on a moving and fixed ring (Fig. 3B) [5]–[8]. The RSR is designed to meet the clinical and mechanical requirements for femur fracture surgery, including 1) inserting traction forces/torques, 2) precise alignment, and 3) holding bone fragments for fixation [5]–[8]. To represent RSR in the unreal engine, the robot components are divided into varying links ($L_{RSR}$1-5) and connected via joints ($J_{RSR}.A_i$-$D_i$) using the parent-child convention to define the relationship between the links (Fig. 3B). Each arm of the Robossis surgical robot includes three joints: universal (represented as an active and passive joint ($J_{RSR}.A_i$ & $J_{RSR}.B_i$)), prismatic ($J_{RSR}.C_i$), and spherical ($J_{RSR}.D_i$) (Fig. 3B). The universal joint ($L_{RSR}2_i$) connects the rotary actuator shaft ($L_{RSR}1$) to the lower arm ($L_{RSR}3_i$) and is placed in the fixed platform. Also, the spherical joint connects the upper parts of the linear actuators ($L_{RSR}4_i$) to the moving ring ($L_{RSR}5_i$). As such, a closed-loop link-joint relationship is defined in the RSS that resembles the real-world equivalence of the RSR.

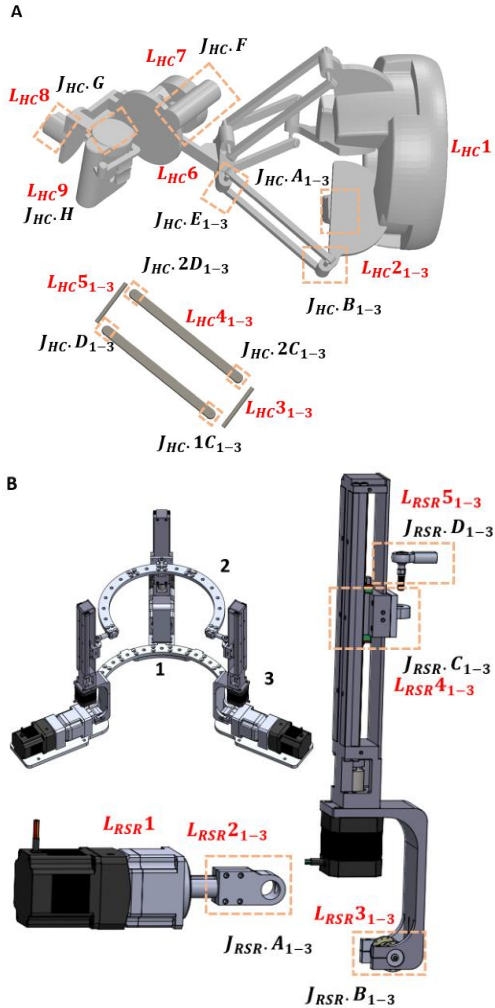

Figure 3. Kinematic representation of the leader-follower Robossis system within unreal engine. A & B) The HC Sigma.7 and RSR strucutre is divided into vary links and connected via joints to define a closed-loop link-joint relationship that resemble the actual real-world. B) RSR structure include a fixed ring (1), a moving ring (2), and three arms (3) where each arm consists of a linear and rotary actuator.

### C. 2D Fluoroscopic Imaging

We developed a 2D fluoroscopic imaging within RSS to enable various anatomical planar views of the surgical field. The 2D fluoroscopic imaging was developed by leveraging the scene capture 2D tool within the unreal engine. The scene capture 2D setting is specified to show only components within the surgical setting, including the patient's thigh, proximal and distal bone, and the RSR. Furthermore, the material properties of the patient thigh and bone were optimized to create an x-ray-like effect. Specifically, the material properties were set to be as an additive blended mode while specifying the opacity for the bone and thigh materials as 0.8 and 0.1, respectively. Also, the scene capture 2D was set as a child of the C-arm static mesh to receive input from the Meta Quest controller for the user-desired global rotation.

## V. HAPTIC CONTROLLER AND MOTION CONTROL

### A. Motion Control

We drive the motion of the leader Sigma.7 HC active joints ($\theta_{HC,i}$, labeled in Fig3. A as $J_{HC}A_{1-3}$, $J_{HC}F$, $J_{HC}G$, and $J_{HC}H$) and the follower RSR active joints ($\theta_{RSR,i}$, labeled in Fig3. B as $J_{RSR}A_{1-3}$) and linear actuators ($d_{RSR,i}$, labeled in Fig3. B as $J_{RSR}C_{1-3}$) within the Unreal Engine to recreate a realistic real-world movement. As described earlier, the Sigma.7 HC is a hybrid structure composed of a delta mechanism with three active joints ($\theta_{HC,1-3}$) and a wrist mechanism with three active joints ($\theta_{HC,4-6}$). We determine the active joint angles ($\theta_{HC,1-6}$) from the HC (Force Dimension SDK) library. As such, we interface the active joint values into the RSS-designed blueprint to drive the HC end-effector into the theoretical global position and orientation.

Furthermore, the HC Sigma-7 end-effector global position and orientation trajectories, as commanded by the user's hand, are interfaced with the RSR as an incremental trajectory as

$$x(t)_{RSR} = x_{RSR}(t-1) + (x(t)_{HC} - x(t-1)_{HC}) * S \quad (1)$$

where $x(t)_{RSR} \in R^6$ and $x_{RSR}(t-1) \in R^6$ are the current and previous location of the RSR, and $x(t)_{HC} \in R^6$ and $x(t-1)_{HC} \in R^6$ are the current and previous location of the HC (user's hands). Also, $S \in R^2$ is the dynamic scaling factor and defined as

$$S = \frac{[Max_v, \; Max_\omega]}{[\|v_{HC}\|, \; \|\omega_{HC}\|]} \quad (2)$$

where $\|v_{HC}\|$ and $\|\omega_{HC}\|$ are the norms of the linear and angular velocities of the HC (user's hands) during motion ~ $\dot{x}(t)_{HC} \in R^6$. Also, $Max_v$, $Max_\omega$ are the desired maximum linear and angular velocities based on the user's desired input.

Further, we map the input of the user's hand scaled trajectory's location and orientation ($x(t)_{RSR}$) as the desired location of the Robossis end effector (center of the moving ring ($P$)). Given the position (P(x, y, z)) and orientation (R (α, β, γ)) of the endpoint effector ($P$), the length of the linear actuator ($d_{RSR,i}$) and the rotation of the active joint ($\theta_{RSR,i}$) are computed as derived in our previous work [5]–[8].

Given the desired position of the linear actuator ($d_{RSR,i}$), active joints angle of the RSR ($\theta_{RSR,i}$) and HC ($\theta_{HC,i}$), we specify the angular drive parameter for each joint within the Unreal Engine to define the physical strength of the joints (stiffness, damping, and maximum force limit). Algorithm 1 below describes the overall procedure used to drive the HC, and RSR in the RSS simulator. The algorithm is a blueprint C++ inherited class designed for the RSS to interface the HC and motion control of the RSR and HC.

**Algorithm 1: Motion Control**

1: MotionControl::BeginPlay () {
2:   ActiveJoint_i->SetAngularDriveParams(stiffness, damping, force)
3:   ActiveJoint_i->SetAngularOrientationDrive(true, true)
4:   LinearJoint_i->SetLinearDriveParams(stiffness, damping, force)
5:   LinearJoint_i->SetLinearPositionDrive(true,true,true)
6: }
7: MotionControl::Tick(DeltaTime) {
8:   $[x(t)_{HC}]$ = Sigma_7 → GetPositionRotation ()
9:   $[\dot{x}(t)_{HC}]$ = Sigma_7 → GetLinearAngularSpeed ()
10:  $[\|v_{HC}\|, \|\omega_{HC}\|]$ → norm($\dot{x}(t)_{HC}$)
11:  If ($\|v_{HC}\|, \|\omega_{HC}\| > Max_v, Max_\omega$) {
12:      $S$ → $Eq.(2)$ }
13:  Else {
14:      $S$ = [1.0, 1.0]
15:  }
16:  $x(t)_{RSR}$ → $Eq.(1)$
17:  Robossis_Kinematics($x(t)_{RSR}$)
18:  SetLinearPositionTarget($d_{RSR,i}$)
19:  SetAngularOrientationTarget($\theta_{RSR,i}$)
20:  $[\theta_{HC,i}]$ = Sigma_7 → GetJointAngles()
21:  SetAngularOrientationTarget($\theta_{HC,i}$)
22: }

*B. Haptic Feedback: Bone Collision*

We develop a haptic feedback bone collision algorithm to prevent the user from overlapping the distal (D) and proximal (P) bone (D ∩ P). We model the proximal bone as a fixed oriented bounding box (OBB) while the distal bone is modeled as a moving OBB with respect to the center of the moving ring of the RSR (Fig. 4). Therefore, 4 OBBs are designed to cover the shaft, distal, and proximal segments of the femur bone for realistic modeling.

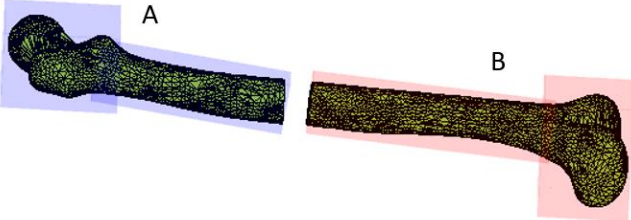

Figure 4. Oriented bounding boxes (OBB) for the proximal bone (A) and distal bone (B) model is illustrated. A total of 4 OBBs are designed to cover the shaft, distal, and proximal segments of the femur bone for a realistic modeling.

We implement the separating axis theorem (SAT) to detect the collision between the proximal and distal bone OBBs. As such, we develop the collision algorithm to check if there is an overlap between the proximal and distal OBBs for each potential separating axes ($L$) that include the 3-faces normal for each of the OBBs and the additional 9 potential separating axis arising from the cross products between the edges of OBBs. Therefore, there are 15 possible separating axes that we need to verify to determine if there is a collision occurring between 2 OBBs, one from the proximal end and one from the distal end. Since the proposed model consists of 2 proximal OBBs and two distal OBBs, we check if there is a collision between each distal OBB with respect to the proximal OBB. Given each potential separating axis ($L$), the projection of the OBBs extent into the potential separating axes is estimated as:

$$E = W_x|proj_L A_x| + W_y|proj_L A_y| + W_z|proj_L A_z| \quad (3)$$

where $W_x$, $W_y$, and $W_z$ are the half length of the OBBs faces, $L$ is the potential separating axes, and $A_x, A_y,$ and $A_z$ are the axis of each of the local faces of the OBBs. Furthermore, the maximum and minimum extent of the OBBs that is projected into the potential separating axes $(L)$ can be estimated as:

$$BE = proj_L C \pm E \quad (4)$$

Where $C$ is the local XYZ center of the OBBs. As such, we can determine if there is an overlap (OL) between the OBBs as:

$$OL = min(\text{maxBE}_{OBB,P}, \text{maxBE}_{OBB,D}) - max(\text{minBE}_{OBB,P}, \text{minBE}_{OBB,D}) \quad (5)$$

Where OL is the overlap between the maximum and minimum extent of the OBBs from each distal and proximal segment. Given the iteration for each potential separating axis, OL < 0 indicates the presence of a separating axis; therefore, a collision is not present. On the other hand, if OL > 0 for each potential separating axis ($L$), a collision is present. As such, the force restriction that prevents the motion of the user's hand from overlapping the proximal and distal bone during the simulation is modeled as:

$$F_{Col} = d \cdot \hat{n} \cdot k \quad (6)$$

where $F_{col}$ is the force vector at the contact of the collision, d is the penetration depth, $\hat{n}$ is the norm of the force, and k is the spring constant (1000 N/m). OL with the smallest overlap corresponds to the penetration depth (d) with a normal vector corresponding to the face of the collision ($\hat{n}$). Further, the direction of the norm ($\hat{n}$) is assigned based on the alignment of the vector originating form the center of the colliding distal OBB to the proximal OBB projected on the potential separating axis ($L$). Thus, $F_{Col}$ is the sum due to the collision of each distal OBB with respect to the proximal OBB. Hence, the global force is estimated as:

$$F_g = F_{Col} - v * c \quad (7)$$

where $v$ is velocity vectors and c is the damping constant (10 N s/m). An illustration of the haptic feedback bone collision method is presented in algorithm 2.

## VI. SIMULATION AND TESTING

*A. Robossis Kinematic Interface*

The deviation of the RSR from the motion of the user's hand via the Sigma-7 HC was evaluated. As the simulation proceeded, the user simultaneously manipulated the RSR in all 6-DOF (translational and rotational). Fig. 5A & B present the corresponding trajectories from the RSR (left) and HC (right). We performed an error analysis to determine the deviation of the RSR from the HC (Fig. 5C). Fig. 5C illustrates a maximum variation for translation and rotation as ~ 5 mm and ~ 0.6 deg, respectively.

*B. Haptic Feedback*

The haptic feedback bone collision algorithm is implemented to recreate a realistic scenario in the real physical world. The modeling of OBBs was required due to the curved structure of the femur bone and the 6-DOF movement of the distal bone with respect to the center of the moving ring of the RSR. Simulation analysis illustrated in Fig. 6 A & B shows the force vector relative to the colliding surfaces of the distal bone (red). Fig. 6 A shows the collision of the bone segments when the distal end is aligned with the global XYZ axis, whereas Fig. 6 B shows the collision of the distal end as rotated with respect to the global XYZ axis. In each presented scenario, the force vector is normal to the distal end of the colliding surface of the bone.

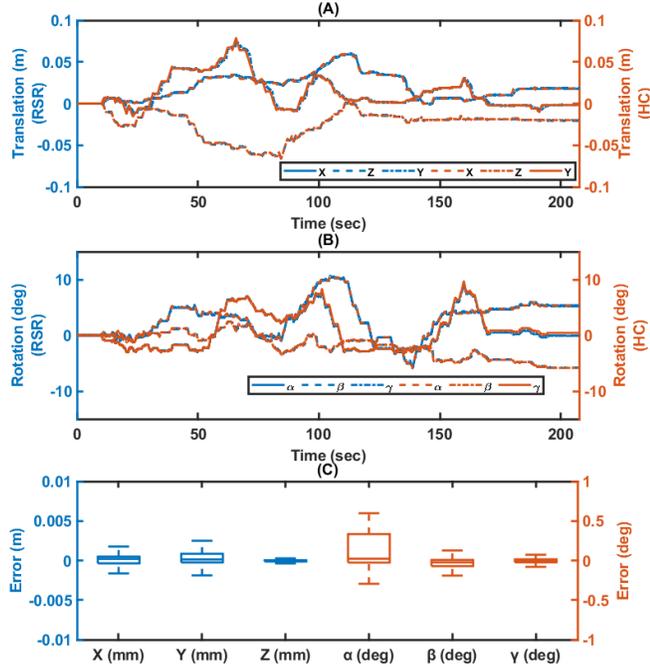

Figure 5. A & B) RSR trajectory as commanded by the user's hand via the HC Sigma-7. C) The corresponding error analysis is performed to determine the deviation of the RSR from the HC.

---

Algorithm 2: Haptic feedback bone collision

---

1: For (1 to 2) { \\ Each distal OOB
2:   For (1 to 2) { \\ Each proximal OBB
3:     $C_{OBB,P}$, $C_{OBB,D}$ → proximal and distal center local (XYZ) position
4:     $A_{x,OBB,P}$, $A_{y,OBB,P}$, $A_{z,OBB,P}$ → local faces axis of OBB P ($R_P^{3X3}$)
5:     $A_{x,OBB,D}$, $A_{y,OBB,D}$, $A_{z,OBB,D}$ → local faces axis of OBB D ($R_D^{3X3}$)
6:     SmallestOverlap → inf
7:     // Check OBBs faces and edges
8:     $AXIS$ = [$R_P^{3X3}$, $R_D^{3X3}$, cross($R_P^{3X3}(i,:)$, $R_D^{3X3}(j,:)$)] // i & j (1 to 3)
9:     For (i = 1 to 15) // for each potential separating axis
10:       L = $AXIS$(:,i)
11:       $E_{OBB,P}$, $E_{OBB,D}$ → Eq. 3
12:       $BE_{OBB,P}$, $BE_{OBB,D}$ → Eq. 4
13:       OL → Eq. 5
14:       If (OL < 0) // no separating axis
15:         d = 0
16:         $\hat{n}$ = [0 0 0]
17:         return
18:       Else // a collision detected
19:         If (OL < SmallestOverlap)
20:           SmallestOverlap → d
21:           If (($C_{OBB,D} - C_{OBB,P}$) · $L > 0$) \\ direction of the norm
22:             $\hat{n}$ → $L$
23:           Else
24:             $\hat{n}$ → -$L$
25:     $F_{Col}$ += Eq. 6 (sum forces)
26: $F_g$ → Eq. 7
27: Sigma_7 → SetForce([$F_g(x,y,z)$]

### C. Virtual Reality Simulation Testing

The RSS environment was created to immerse the trained users in a realistic operating room environment for femur fracture surgery using the Robossis system (Fig. 7). To interact with the environment, the HC Sigma-7 was used to manipulate the distal bone segment in the desired translational and rotational directions (attached video). As the user manipulates the HC, real-time visual rendering for the location of the bone is displayed as 2D fluoroscopic imaging via the Meta Quest headset. Utilizing the Meta Quest controller, the user is able to rotate the c-arm to the desired anatomical planar views. Also, with the implementation of the HC, the user is prevented from overlapping the moving distal bone (attached to the RSR) with the proximal bone to recreate a realistic real-life condition.

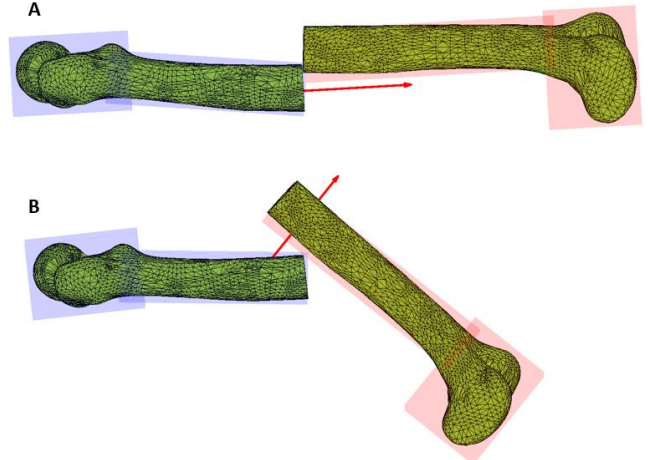

Figure 6. Haptic feedback of the bone collision algorithm between the proximal and distal bone segments is illustrated. A) Shows the collision of the bone segments when the distal end is aligned with the global XYZ axis, whereas B) shows the collision of the distal end as rotated with respect to the global XYZ axis. The force vector, depicted in red, is normal to the colliding surface with respect to the distal bone.

## VII. DISCUSSION

The RSS is designed with the goal of immersing the user in a realistic environment, as previously completed in a cadaveric study. Therefore, surgeons and operating staff will better translate their techniques to the real world. This replication allows the surgeons to gain a more spatial and visual feel that eliminates adjustments needed for the real-world transition. Also, the RSS is designed to provide future trainees with the necessary tools to enhance surgical efficacy for the integration of the RSR in the clinical field.

Further, the proposed methods for the development of the RSS present a novel approach for the representation of digital robots and integration with real-world systems. Specifically, the kinematic representation and matching between the real-world HC Sigma-7, RSR, and the virtual RSR and HC yields real-time evaluation during surgical training. This kinematic matching ensures that the RSR follows the desired motion as the surgeon manipulates the HC.

Additionally, the integration of haptic feedback into the RSS provides users with the virtual representation and collision of the bone segments during the training. Thus, realistic behavior is experienced during training on the simulator. Also, the development of the haptic collision algorithm was required due to the unstable behavior of the Unreal Engine collision detection algorithm known as "SweepMultiByChannel." While attempting to implement the "SweepMultiByChannel," many challenges were faced, including the jumping of the penetration depth and norm, which resulted in an undesired haptic feedback behavior.

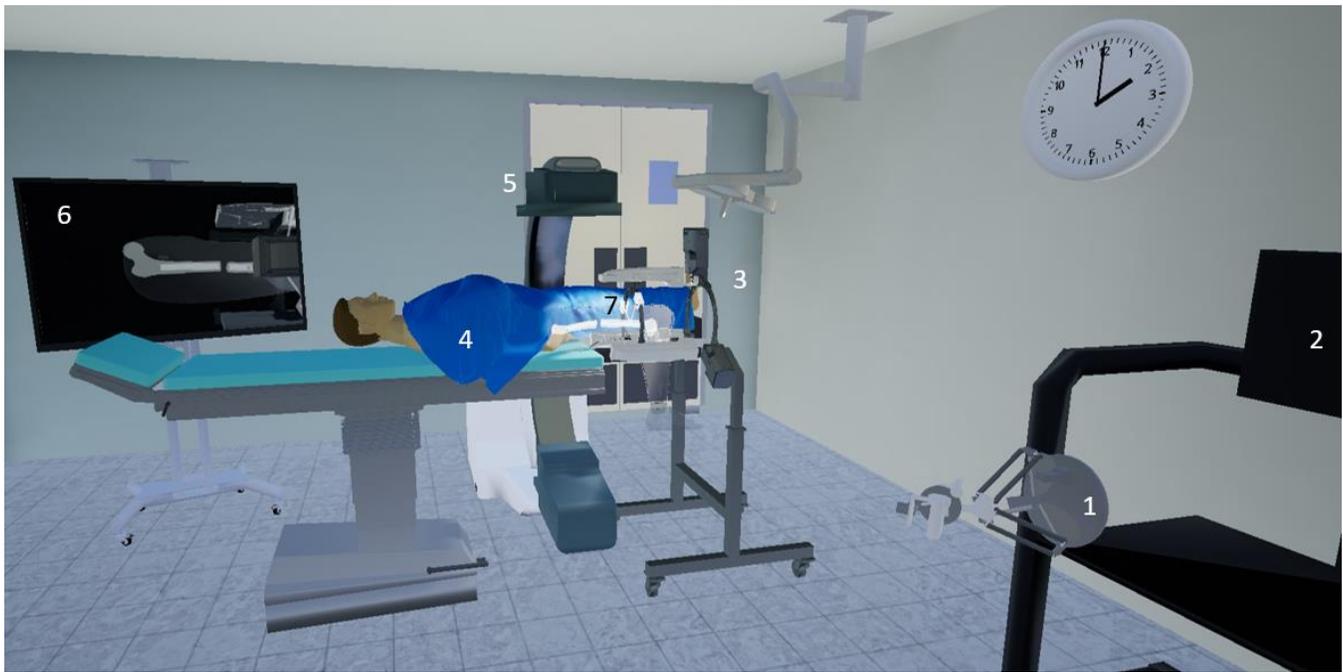

Figure 7. Robossis-assisted femur fracture surgery environment was designed to include (1) a haptic controller, (2) a surgeon workstation, (3) the Robossis surgical robot, (4) a cadaver patient, (5) a C-arm X-ray machine, and (6) real-time visual rendering for the location of the bone is displayed as a 2D fluoroscopic imaging. Robossis surgical robot is attached to the distal bone segment using surgical rods (7). The environment was created with the goal to immerse the trained users in a similar operating room environment for femur fracture surgery using the Robossis system.

For future work, we plan on creating a lower extremity muscle model using the Hill-based Model to insert forces onto the user's hands during the simulation. This will be necessary to develop the operational experience surgeons need for femur fracture reduction surgery. Also, we plan on testing the RSS with surgeons and operating staff to get feedback on the experience-based need for the Robossis system.